# A Systematic Review for Transformer-based Long-term Series Forecasting

Liyilei Su[1,2,3#], Xumin Zuo[1,3#], Rui Li[3], Xin Wang[3], Heng Zhao[3*], and Bingding Huang[1,3*]

[1]College of Applied Sciences, Shenzhen University, Shenzhen, 518060, China
[2]Guangdong Key Laboratory for Biomedical Measurements and Ultrasound Imaging, National-Regional Key Technology Engineering Laboratory for Medical Ultrasound, School of Biomedical Engineering, Shenzhen University Medical School, Shenzhen 518060, China
[3]College of Big Data and Internet, Shenzhen Technology University, Shenzhen, 518188, China
#These authors contributed equally to this work.
*Corresponding authors: Heng Zhao (zhaoheng@sztu.edu.cn) and Bingding Huang (huangbingding@sztu.edu.cn)

**Abstract**

The emergence of deep learning has yielded noteworthy advancements in time series forecasting (TSF). Transformer architectures, in particular, have witnessed broad utilization and adoption in TSF tasks. Transformers have proven to be the most successful solution to extract the semantic correlations among the elements within a long sequence. Various variants have enabled transformer architecture to effectively handle long-term time series forecasting (LTSF) tasks. In this article, we first present a comprehensive overview of transformer architectures and their subsequent enhancements developed to address various LTSF tasks. Then, we summarize the publicly available LTSF datasets and relevant evaluation metrics. Furthermore, we provide valuable insights into the best practices and techniques for effectively training transformers in the context of time-series analysis. Lastly, we propose potential research directions in this rapidly evolving field.

## 1 Introduction

The time series is usually a set of random variables observed and recorded sequentially over time. Key research directions for time-series data are classification [1, 2], anomaly detection [3-5], event prediction [6-8], and time series forecasting [9-11]. Time series forecasting (TSF) predicts the future trend changes of time series from a large amount of data in various fields. With the development of data collection technology, the task gradually evolves into using more historical data to predict the longer-term future, which is long-term time series forecasting (LTSF) [12, 13]. Precise LTSF can offer support to decision makers to better plan for the future by forecasting outcomes further in advance, including meteorology prediction [14], noise cancellation [15], financial long-term strategic guidance [16], power load forecasting [17, 18], and traffic road condition prediction [19].

Formerly, traditional statistical approaches were applied to time series forecasting, such as autoregressive (AR) [20], moving average (MA) [21] models, auto-regressive moving average (ARMA) [22], AR Integrated MA (ARIMA) [23], and spectral analysis techniques [24]. However, these traditional statistical methods require many a priori assumptions on the time-series prediction, such as stability, normal distribution, linear correlation, and independence. For example, AR, MA, and ARMA models are based on the assumption that time series are stationary, but in many real cases, time-series data exhibit non-stationarity. These assumptions limit the effectiveness of these traditional methods in real-world applications.

As it is difficult to effectively capture the nonlinear relationships between time series with traditional statistical approaches, many researchers have studied LTSF from the perspective of machine learning (ML) [25-29]. Support vector machines (SVMs) [30] and adaptive boosting (AdaBoost) [31] were employed in the field of TSF. They calculate data metrics, such as minimum, maximum, mean, and variance, within a sliding window as new features for prediction. These models have somewhat solved the problem of predicting multivariate, heteroskedastic time series with nonlinear relationships. However, they suffer from poor generalization, which leads to limited prediction accuracy.

Deep learning (DL) models (Fig.1) have greatly improved the nonlinear modeling capabilities of TSF in recent years. These models are constructed with neural network structures with powerful nonlinear modeling capabilities to learn complex patterns and feature representations in time series automatically. Therefore, DL is an effective solution for TSF and many other problems related to TSF, such as hierarchical time series forecasting [32], intermittent time series forecasting [33], and sparse multivariate time series forecasting [34] asynchronous time series forecasting [35, 36]. It has even extended some multi-objective, multi-granular forecasting scenarios [37] and multi-modal time series forecasting scenarios [38, 39]. The advantage of deep learning models can be attributed to their profound flexibility and ability to capture long-term dependencies and handle large-scale data.

It is noteworthy that recurrent neural networks (RNNs) [40] and their variants, such as long short-term memory networks (LSTMs) [41] and gated recurrent units (GRUs) [42-44], are widely employed among deep learning models to process sequence data. These models process batches of data sequentially using a gradient descent algorithm to optimize the unknown model parameters. The gradient information of the model parameters is updated by back-propagation through time [45]. However, due to the sequential processing of input data and back-propagation through time, they suffer from some limitations, especially when dealing with datasets with long dependencies. The training process of LSTM and GRU models also suffers from gradient vanishing and explosion. Though some architectural modifications and training techniques can help LSTM and GRU to alleviate the gradient-related problems to some extent, the effectiveness and efficiency of RNN-based models may still be compromised [46]. Furthermore, it is possible to apply models like Convolutional Neural Network (CNN) to conduct time-series analysis.

On the other hand, the transformer [47] is a model that combines various mechanisms, such as attention, embedding, and encoder-decoder structures, in natural language processing. Later, studies improved the transformer and gradually applied it to TSF, imaging, and other areas, making transformers progressively their genre. Recent advancements in transformer-based models have shown substantial progress [12, 48, 49]. The self-attentive mechanism of the transformer allows for adaptive learning of short-term and long-term dependencies through pairwise (query-key) request interactions. This feature grants the transformer a significant advantage in learning long-term dependencies on sequential data, enabling the creation of more robust and expansive models [50]. The performance of transformers on LTSF is impressive, and they have gradually become the current mainstream approach.

The two main tasks of time-series data are forecasting and classification. Forecasting aims to predict real values from given time-series data, while the classification task categorizes given time-series data into one or more target categories. Many advances have been made in time-series Transformers for forecasting [12, 49, 51-59] and classification tasks [1, 60-62]. However, genuine time-series data tends to be noisy and non-stationary, and the learning of spurious dependencies, lacking interpretability, can occur if time-series-related knowledge is not combined. Thus, challenges remain despite the notable achievements in accurate long-term forecasting using transformer-based models.

In this review, we commence with a comprehensive overview of transformer architecture in Section 2. Section 3 presents transformer-based architectures for LTSF in recent research. In Section 4, we analyze transformer effectiveness for LTSF. Subsequently, Section 5 summarizes the public datasets and evaluation metrics in LTSF tasks. Section 6 introduces several training strategies in existing transformer-based LTSF solutions. Finally, we conclude this review in Section 7.

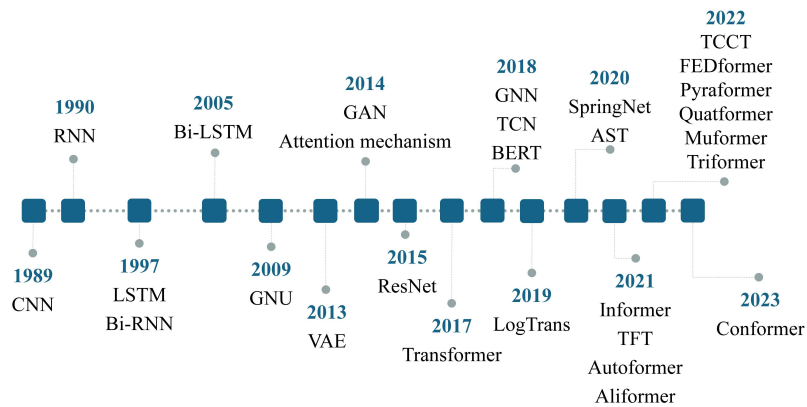

Fig.1 The development history of TSF algorithms based on deep learning.

## 2 Transformer

In this section, we begin by analyzing the inherent mechanics of the transformer proposed by Vaswani et al. [63] in 2017, with the objective of presenting solutions to the challenge of neural machine translation. Fig.2 shows the transformer architecture. Subsequently, we delve into the operations within each constituent of the transformer and the underlying principles that inform these operations. Several variants of the transformer architecture have been proposed for time-series analysis; however, our discussion in this section is limited to the original architecture [64] [65].

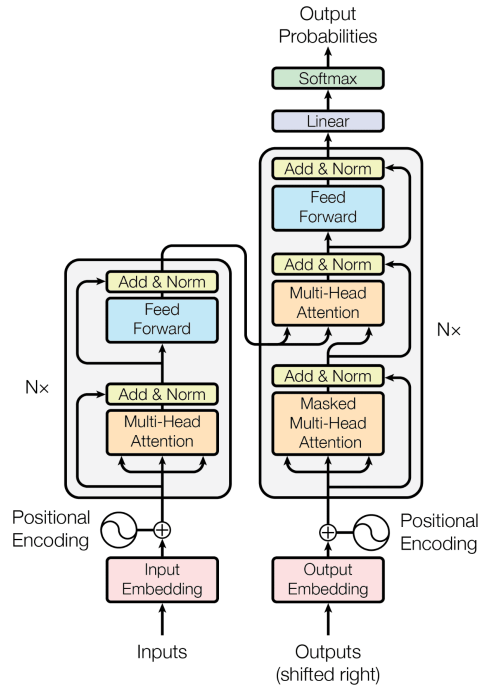

Fig.2 Schematic diagram of transformer

## 2.1 Self-attention

The self-attention mechanism is a process that involves mapping a query and a sequence of key-value pairs to generate a corresponding output vector. The resulting vector is determined by the summation of weights acting on values computed from the query and key. A schematic representation of the self-attention mechanism is depicted in Fig.2.

Scaled Dot-Product Attention

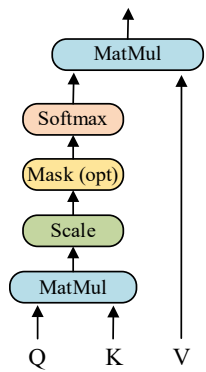

Fig.3 Schematic diagram of self-attention

As can be seen in Fig.3, the core of the self-attention mechanism is to get the attention weights by calculating Q and K and then acting on V to get the whole weights and outputs. Q, K, and V are the Query, Key, and Value matrices of the input sequence after linear transformation. With respect to the input sequence denoted as X, the parameters Q, K, and V are given by

$$Q = W_q X, K = W_k X, \text{and } V = W_v. \tag{1}$$

Q, K, and V are computed by multiplying the input X by three different matrices (but this is only limited to the encoder and decoder encoding process using the self-attention mechanism in their respective inputs; the Q, K, and V in the interaction between the encoder and decoder are referred to otherwise). Here, the computed Q, K, and V can be interpreted as three different linear transformations of the same input to represent its three different states. After Q, K, and V are computed, the weight vectors can be further computed. Specifically, for the inputs Q, K, and V, the weight vectors are calculated as:

$$\text{Attention}(Q, K, V) = \text{softmax}\left(\frac{QK^T}{\sqrt{d_k}}\right) V. \qquad (2)$$

The dimension of the query and key is denoted by $d_k$. The attention for each position is normalized using the softmax function. The formula illustrates that the attention score matrix can be derived by executing a dot product operation between the query and key, followed by division with a scaling factor of $\sqrt{d_k}$. Subsequently, the attention weights for each position are obtained by performing a softmax operation on the attention score matrix. The ultimate self-attention representation is achieved by multiplying the attention weights with the value matrix. The compatibility function employed in this process is a scaled dot product, thus rendering the computation process efficient. Additionally, the linear transformation of the inputs introduces ample expressive power. As illustrated in Fig. 2, the scale process corresponds to the division of $d_k$ in Eq. 2. It is imperative to note that scaling is essential because, for larger $d_k$, the value obtained after $QK^T$ is excessively large, consequently causing a diminutive gradient after the softmax operation. The diminutive gradient hinders the training of the network and, thus, is not conducive to the overall outcome.

## 2.2 Multi-Head Attention

The self-attention mechanism serves as a solution to the sequential encoding challenge encountered in conventional sequence models. It enables the generation of a final encoded vector that incorporates attention information from multiple positions, achieved through a finite number of matrix transformations on the initial inputs. However, it is worth noting that the model's encoding of positional information may lead to an overemphasis on its own position, potentially neglecting the importance of other positions. To address this issue, the Multi-Head Attention mechanism has been introduced.

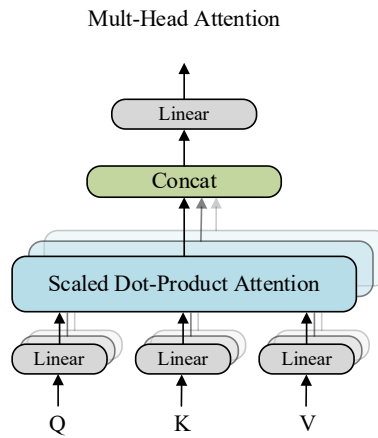

Fig.4 Multi-head attention

As shown in Fig.4, the multi-attention mechanism is a self-attention processing process of multiple groups of the original input sequences; and then each group of self-attention results are spliced together to perform a linear transformation to obtain the final output results. Specifically, its

calculation formula is:

$$\text{Multi-Head}(Q, K, V)\text{Concat} = (\text{head}_1, \dots, \text{head}_h)W_O$$
$$\text{head}_i = \text{Attention}(QW_{Qi}, KW_{Ki}, VW_{Vi}). \quad (3)$$

In this context, the matrices Q, K, and V refer to the query, key, and value matrices of the input sequences, respectively, subsequent to linear transformation. The variable h denotes the number of attention heads. Additionally, the weight matrices $W_{Qi}, W_{Ki}, and W_{Vi}$ are utilized to carry out the linear transformation on Q, K, and V. The output weight matrix of the multi-head attention is denoted by the symbol $W_O$. The computation of a single attention head is denoted by attention in Eq. 2, equivalent to the previously mentioned self-attention mechanism. Each attention head maps the inputs through independent linear transformations and subsequently applies the attention mechanism to obtain the representation. The final output of the multi-head attention is obtained by combining the representations of all attention heads and applying a linear transformation to the output weight matrix $W_O$.

## 2.3 Encoder and decoder

The encoder is depicted on the left-hand side of Figure 2. The encoder component encompasses two primary networks: the multi-head attention mechanism and the two-layer feed-forward neural network. Notably, residual connections are added to both network parts, and layer normalization is implemented after residual connections. Consequently, each part's output is represented as $LayerNorm(x + Sublayer(x))$, with the Dropout operation added to both. For the two-layer, fully connected network in the second part, the specific computational procedure is outlined as follows.

$$\text{FFN}(x) = \max(0, xW_1 + b_1)W_2 + b_2 \quad (4)$$

The variable x symbolizes the feature representation of the input. $W_1$ and $W_2$ denote the weight matrices, while $b_1$ and $b_2$ represent the bias vectors. The $\max(0, \dots)$ signifies the utilization of the Rectified Linear Unit (ReLU) as the activation function.

The decoder is similar to the Encoder, albeit with the inclusion of an extra multi-head attention mechanism that interacts with the encoder output. Unlike the encoder, the decoder comprises three parts of the network structure. The top and bottom segments resemble the encoder, save for a middle section that engages with the encoder's output (Memory), referred to as "encoder-decoder attention." In this component, the input Q is derived from the output of the following multi-head attention mechanism, while K and V are linearly transformed outputs (Memory) of the encoder component.

## 2.4 Positional encoding

In the context of modeling text-related data, the initial step involves its vectorization. In machine learning, one-hot coding, bag-of-words models, and TF-IDF are frequently employed techniques to represent text. However, in deep learning, it is more common to associate individual words (or tokens) with a low-dimensional, dense vector space using an embedding layer. Consequently, in the transformer model, the first task is also to vectorize the text in this manner and name it token embedding, which is commonly referred to as word embedding in deep learning.

Adopting a prior network model, such as CNN or RNN, would signify the conclusion of text vectorization, as these network architectures already possess the capacity to capture temporal features, whether in the n-gram form in CNNs or the temporal form in RNNs. However, this does not apply to the transformer network architecture, which eliminates recursion and convolution. The self-attention mechanism's introduction principle indicates that the actual operation of the self-attention mechanism is solely a linear transformation accomplished by multiplying multiple matrices back and forth. Therefore, even when the word order is disrupted, the result remains unaltered; in other words, the original text sequence will be lost solely if the self-attention mechanism is utilized.

As illustrated in Fig.5, the sequence "I am writing review" underwent a linear transformation after the word embedding representation. Subsequently, we modified the sequence to "review am writing I" and executed a linear transformation employing the same weight matrix in the intermediary. Based on the calculation outcomes depicted in the figure, no fundamental disparity exists between the results prior to and after the exchange of sequence positions, with only the corresponding positions being interchanged. Considering these issues, the transformer adds an additional positional embedding after the token embedding of the original input text to delineate the temporal sequence of the data.

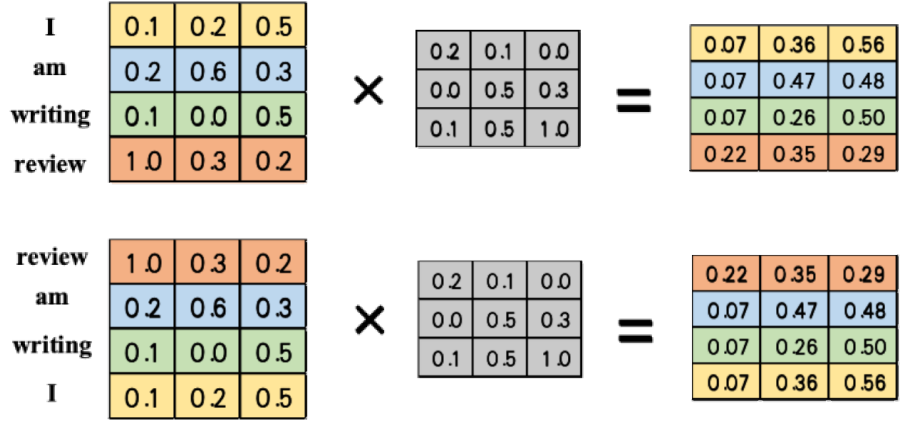

Fig.5 Word embedding matrix for the sequence "I am writing review"

In the original transformer, the author employs the sinusoidal functions to generate positional information for each dimension.

$$\text{PosEmb}(pos, 2i) = \sin\left(\frac{pos}{10000^{2i/d_{\text{model}}}}\right) \tag{5}$$

$$\text{PosEmb}(pos, 2i+1) = \cos\left(\frac{pos}{10000^{2i/d_{\text{model}}}}\right) \tag{6}$$

The present study introduces the concept of $\text{PosEmb}(pos, 2i)$ as a representation of position encoding for position *pos* and even index 2i. Similarly, $\text{PosEmb}(pos, 2i+1)$ signifies the position encoding of position *pos* and odd index 2i+1. Notably, $d_{\text{model}}$ refers to the dimensions of a sequence in the transformer model. Incorporating this non-linear positional embedding position information contributes to the comparison results in Fig.6.

Fig.6 reveals that the outcomes after a linear conversion utilizing an identical weight matrix display notable disparities pre- and post-rearrangement. Thus, it can be inferred from Fig.6 that the insufficiency of the self-attentive mechanism in capturing temporal sequencing information can be compensated for by applying positional embedding.

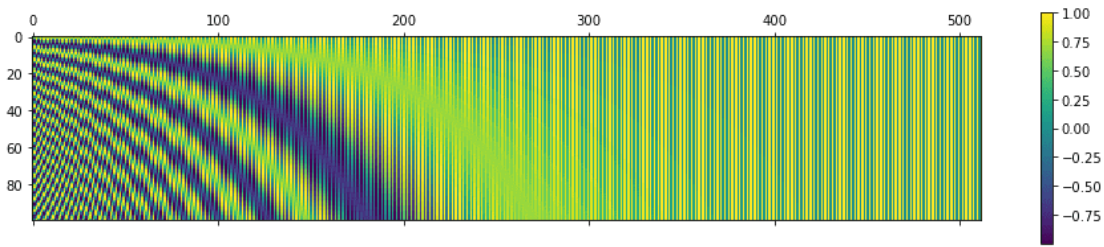

Fig.6 Positional embedding matrix for the sequence "I am writing review"

Fig.7 depicts a graphical representation of the position matrix. The vertical axis denotes each Token in the input sequence, and each row corresponds to the embedding vector $\vec{P}$. Based on the assumptions outlined in the original transformer, the sequence length is 100, and the dimension is 512. Each row contains 512 values—each with a value between 1 and –1. Consequently, one can infer that the vectors' positions on any two tokens can be interchanged, and an information model that incorporates positional cues, as illustrated in the accompanying figure, can indeed detect this distinction.

Fig.7 The 512-dimensional positional embedding for a sentence with a maximum length of 100. Each row represents the embedding vector $\vec{P}$

**3 Transformer-based architectures in LTSF**

The design of a network needs to consider the characteristics and nature of problems. In this section, we first analyze the key problems in the LTSF tasks, followed by a discussion of some popular recent transformer-based architectures in LTSF tasks.

**3.1 LTSF's key problems**

LTSF is usually defined as forecasting a more distant future [12, 66]. Given the current status of the existing work, there are two main problems in the field of LTSF: complexity and dependency. LTSF requires processing a large amount of data [67], which may lead to longer training times and require more computational resources [68], as the computational complexity grows exponentially with the length of the sequence. Additionally, storing the entire sequence in memory may be challenging due to the computer's limited memory [69]. This may limit the length of the time series available for prediction [70].

Meanwhile, LTSF models need to have the ability to accurately capture the temporal relationship between past and future observations in a time series [71-73]. Long-sequence time series exhibit long-term dependence [74, 75], challenging the models' ability to capture dependencies [12]. Moreover, LTSF is characterized by inherent periodicity and non-stationarity [76], and thus, LTSF models need to learn a mixture of short-term and long-term repeated patterns in a given time series [67]. A practical model should capture both repeated ways to make accurate predictions, which imposes more stringent requirements on the prediction model regarding learning dependence.

**3.2 Transformer variants**

A transformer [47] mainly captures correlations among sequence data through a self-attention mechanism. Compared with the traditional deep learning architecture, the self-attention mechanism in a transformer is more interpretable. Wu et al. [53] introduced the vanilla transformer to the field of temporal prediction for influenza disease prediction. However, as mentioned above, transformers have a large computational complexity, leading to high computational costs. Moreover, the utilization of location information is not apparent, the position embedding in the model embedding process is ineffective, and long-distance information cannot be captured. A brief conclusion of recent transformer-based architectures is given in Table 1.

Table 1. Transformer-based architectures.

| Reference | Technique | Brief information |
|---|---|---|
| Wu et al. [53] | Transformer | Transformer for LTSF on influenza |
| LogSparse transformer [49] | LogSparse self-attention + Transformer | Reduce time complexity by convolutional self-attention |
| AST [56] | GAN + Transformer | Reduce error accumulation by sparse attention with GAN |
| SpringNet [77] | Spring attention + Transformer | Spring attention to repeatable long-term dependency fluctuating patterns |
| Lee et al. [78] | Partial correlation-based attention + series-wise multi-resolution | Improve pair-wise comparisons-based attention disadvantages with partial correlation-based attention |
| Informer [12] | ProbSparse self-attention + self-attention distilling + generative style decoder | Sparse and computationally effective |
| Autoformer [67] | Sequence decomposition + auto-correlation + Transformer | Auto-correlation and sequence decomposition architecture |
| Pyraformer [70] | Pyramidal attention module + Transformer | Multi-resolution representation with pyramidal attention module |
| FEDformer [79] | Fourier enhanced + wavelet enhanced + Transformer | Reduce time complexity with frequency domain decomposition based on Autoformer architecture |
| TCCT [51] | CNN + CSPAttention + Transformer | Reduce computational cost with CSPAttention |
| Chu et al. [80] | Autoformer + Informer + Reformer + MLP | Incorporate multiple transformer variants and meta-learner |
| Quatformer [81] | Learning-to-rotate attention | Quaternion architecture with learning- |

|   | + trend normalization + Transformer | to-rotate attention |
|---|---|---|
| Muformer [68] | Multi-granularity attention + Transformer + Kullback–Leibler | For multi-sensory domain feature enhancement and multi-headed attentional expression enhancement |
| Triformer [13] | Patch Attention + Transformer | Implement the capture of linear complexity and different temporal dynamic patterns of sequences by a triangular, variable-specific attention architecture |
| Conformer [82] | Fourier transform + sliding window + Transformer | Extract correlation features of multivariate variables by fast Fourier transform, and improve the operational efficiency of long-period forecasting with a sliding window approach |

The time complexity of self-attention computation in a transformer is initially established at $O(L^2)$, leading to high computational cost. Some subsequent works have been developed to optimize this time complexity and the long-term dependency of transformer-based models.

The LogSparse transformer [49] model first introduces transformer to the field of TSF, making transformer more feasible for time series with long-term dependencies. LogSparse transformer allows each time step to be consistent with the previous time step and is selected using an exponential step. It proposed convolutional self-attention by employing causal convolutions to produce queries and keys, reducing the memory utilization from $O(L^2)$ to $O(L(logL))$ in each self-attention layer. The prediction accuracy achieved for fine-grained, long-term dependent time series can be improved in cases with limited memory.

Informer [12] uses the ProbSparse self-attention mechanism, further reducing the computational complexity of the traditional transformer model to $O(L(logL))$. At the same time, inspired by dilated convolution in [83] and [84], it also introduced the self-attention distilling operation to remove redundant combinations of value vectors to reduce the total space complexity of the model. In addition, it designed a generative style decoder to produce long sequence outputs with only one forward step to avoid accumulation error. The Informer architecture was tested on various datasets and performed better than models such as Autoregressive Integrated Moving Average (ARIMA) [85], Prophet [86], LSTMa [87], LSTNet [88], and DeepAR [89].

The Autoformer [67] is a simple seasonal trend decomposition architecture with an auto-correlation mechanism working as an attention module. It achieves $O(L(logL))$ computational time complexity. This deep decomposition architecture embeds the sequence decomposition strategy into the encoder-decoder structure as an internal unit of Autoformer.

In contrast, TCCT [51] designs a CSP attention module that merges CSPNet with a self-attentive mechanism and replaces the typical convolutional layer with an expanded causal convolutional layer, thereby modifying the distillation operation employed by Informer to achieve exponential receptive field growth. In addition, the model develops a penetration mechanism for stacking self-attentive blocks to obtain finer information at negligible additional computational costs.

Pyraformer [70] is a novel model based on hierarchical pyramidal attention by letting the maximum length of the signal traversal path be a constant concerning the sequence length L and can achieve theoretical $O(L)$ complexity. Pyraformer conducts both intra-scale and inter-scale attentions, which capture temporal dependencies in an individual resolution and build a multi-resolution representation of the original series, respectively. Similarly, Triformer [13] proposed a triangular, variable-specific attention architecture, which achieves linear complexity through patch attention while proposing a lightweight approach to enable variable-specific model parameters.

FEDformer [79] achieves $O(L)$ linear computational complexity by designing two attention modules that process the attention operation in the frequency domain with the Fourier transform [90] and wavelet transform [91], respectively. Instead of applying transformer to the time domain, it applies it to the frequency domain, which helps it better expose potential periodic information in the input data.

The Conformer [82] model uses the fast Fourier transform to extract correlation features of multivariate variables. It employs a sliding window approach to improve the operational efficiency of long-period forecasting, sacrificing global information extraction and complex sequence modeling capabilities, thereby reducing the time complexity to $O(L)$.

To address the problems of long-term dependency, Lin et al. [77] established SpringNet for solar prediction. They proposed a DTW attention layer to capture the local correlations of time-series data, which helps capture repeatable fluctuation patterns and provide accurate predictions. For the same purpose, Chu et al. [80] combined Autoformer, Informer, and Reformer to propose a prediction model based on stacking ensemble learning.

Chen et al. [81] proposed a Quatformer framework in which learning-to-rotate attention introduces learnable period and phase information to describe complex periodic patterns, trend normalization to model normalization of the sequence representation in the hidden layer, and decoupling of the LRA by using the global memory, to efficiently fit multi-periodic complex patterns in the LTSF while achieving linear complexity without loss of prediction accuracy.

To alleviate the problem of redundant information input in LTSF, the Muformer proposed by Zeng et al. [68] enhances the features by inputting the multi-perceptual domain processing mechanism, while the multi-cornered attention head mechanism and the attention head pruning mechanism enhance the expression of multi-head attention. Each of these efforts takes a different perspective on optimizing the parametric part of the model, but a general architecture and component that can reduce the number of required model parameters has not yet emerged.

In addition to the previously mentioned transformer-based architectures, other architectural modifications have emerged in recent years. For example, the Bidirectional Encoder Representations from Transformers (BERT) [92] model is built by stacking transformer encoder modules and introducing a new training scheme. Pre-training the encoder modules is task-independent, and decoder modules can be added later and fine-tuned to the task. This scheme allows BERT models to be trained on large amounts of unlabeled data. BERT architecture has inspired many new transformer models for time-series data [1, 57, 60]. However, compared to NLP tasks, time-series data include various data types [1, 12, 93]. Thus, the pre-training process will have to be different for each task. This task-dependent pre-training contrasts with the NLP tasks, which can start with the same pre-trained models, assuming all tasks are based on the same language semantics and structure.

Generative adversarial networks (GANs) consist of the generator and the discriminator,

learning adversarially from each other. The generator-discriminator learning principle has been applied to the time-series forecasting task [56]. The authors use a generative adversarial encoder-decoder framework to train a sparse transformer model for time-series forecasting, solving the problem of being unable to predict long series due to error accumulation. The adversarial training process improves the model's robustness and generalization ability by directly shaping the output distribution of the network to avoid error accumulation through one-step-ahead inference.

TranAD [94] applied GAN-style adversarial training with two transformer encoders and two transformer decoders to gain stability. As a simple transformer-based network tends to miss slight deviations of anomaly, an adversarial training procedure can amplify reconstruction errors.

TFT [54] designs a multi-horizon model with static covariate encoders, a gating feature selection module, and a temporal self-attention decoder. It encodes and selects valuable information from various covariates information to perform forecasting. It also preserves interpretability by incorporating global and temporal dependencies and events. SSDNet [95] combines the transformer with state space models (SSM), which use the transformer part to learn the temporal pattern and estimate the SSM parameters; the SSM parts perform the seasonal-trend decomposition to maintain the interpretable ability. While MT-RVAE [96] combines the transformer with Variational AutoEncoder (VAE), it focuses on data with few dimensions or sparse relationships. A multi-scale transformer is designed to extract different levels of global time-series information. AnomalyTrans [60] combines transformer and Gaussian prior association to make rare anomalies more distinguishable. Prior association and series association are modeled simultaneously. The minimax strategy optimizes the anomaly model to constrain the prior and series associations for more distinguishable association discrepancies.

GTA [3] contains a graph convolution structure to model the influence propagation process. Replacing vanilla multi-head attention with a multi-branch attention mechanism combines global-learned attention, multi-head attention, and neighborhood convolution. GTN [62] applies a two-tower transformer, with each tower working on time-step-wise attention and channel-wise attention, respectively. A learnable weighted concatenation is used to merge the features of the two towers. Aliformer [57] makes the time-series sales forecasting using knowledge-guided attention with a branch to revise and denoise the attention map.

In addition, some researchers have made corresponding network improvements for specific applications. First, in the application of transportation，spatiotemporal graph transformer [97] proposes an attention-based graph convolution mechanism for learning a more complex temporal-spatial attention pattern applying to pedestrian trajectory prediction. Traffic transformer [55] designs an encoder-decoder structure using a self-attention module to capture the temporal-temporal dependencies and a graph neural network (GNN) module to capture the spatial dependencies. Spatial-temporal transformer networks introduced a temporal transformer block to capture the temporal dependencies and a spatial transformer block to assist a graph convolution network to capture more spatial-spatial dependencies [98].

There are also applications for event prediction. Event forecasting or prediction aims to predict the times and marks of future events given the history of past events, which is often modeled by temporal point processes (TPP) [6]. Self-attentive Hawkes process (SAHP) [7] and transformer Hawkes process (THP) [8] adopt transformer encoder architecture to summarize the influence of historical events and compute the intensity function for event prediction. They modify the positional encoding by translating time intervals into sinusoidal functions to utilize the intervals between events. Later, a more flexible model named attentive neural datalog through time (ANDTT) [99] was proposed to extend SAHP/THP schemes by embedding all possible events and times with attention.

**4 Transformer effectiveness for LTSF**

Is transformer effective in the time series forecasting domain? The response we provide is affirmative. Since the publication of Zeng's scholarly article, "Are transformers effective for time series forecasting?"[100], the feasibility of utilizing transformer models for time series forecasting has emerged as a significant subject of scholarly discourse. This is particularly noteworthy as a straightforward model emerged victorious over a considerably intricate transformer model, thus prompting a substantial academic discourse. Zeng claimed that the transformer-based models are not effective in time series forecasting. They compare the transformer-based models with a simple linear model, DLinear, which uses the decomposition layer structure in Autoformer and which DLinear claims outperforms the transformer-based models. A transformer with different positional and temporal embeddings retains very limited temporal relationships. It is prone to overfitting on noisy data, whereas a linear model can be modeled in a natural order and with fewer parameters can avoid overfitting. However, Yuqi Nie [101] presents a novel solution to tackle the loss of temporal information induced by the self-attention mechanism. This approach is rooted in the transformer time-series prediction and involves transforming the time-series data into a patch format akin to that of Vision transformer. This conversion preserves the localization of the time series, with each patch serving as the smallest unit for Attention computation. The findings in Table 2 demonstrate that research focused on transformer-based time-series prediction underscores the significance of integrating temporal information to improve the model's prediction performance.

Table 2. Multivariate long-term forecasting results on electricity dataset

| Models | | PatchTST/64 | | PatchTST/42 | | DLinear | | FEDformer | | Autoformer | | Informer | |
|---|---|---|---|---|---|---|---|---|---|---|---|---|---|
| Metric | | MSE | MAE | MSE | MAE | MSE | MAE | MSE | MAE | MSE | MAE | MSE | MAE |
| prediction length | 96 | 0.129 | 0.222 | 0.130 | 0.222 | 0.140 | 0.237 | 0.186 | 0.302 | 0.196 | 0.313 | 0.304 | 0.393 |
| | 192 | 0.147 | 0.240 | 0.148 | 0.240 | 0.153 | 0.249 | 0.197 | 0.311 | 0.211 | 0.324 | 0.327 | 0.417 |
| | 336 | 0.163 | 0.259 | 0.167 | 0.261 | 0.169 | 0.267 | 0.213 | 0.328 | 0.214 | 0.327 | 0.333 | 0.422 |
| | 720 | 0.197 | 0.290 | 0.202 | 0.291 | 0.203 | 0.301 | 0.233 | 0.344 | 0.236 | 0.342 | 0.351 | 0.427 |

On the other hand, the transformer's effectiveness is reflected in Large Language Models. LLMs are powerful transformer-based models, and numerous previous studies have shown that transformer-based models are capable of learning potentially complex relationships among textual sequences [102, 103]. It is reasonable to expect LLMs to have the potential to understand complex dependencies among numeric time series augmented by temporal textual sequences.

The current endeavor for time series LLMs encompasses two primary strategies. One approach involves the creation and preliminary training of a fundamental, comprehensive model specifically tailored for time series. This model can be subsequently fine-tuned to cater to various downstream tasks. This path represents the most rudimentary solution, drawing upon a substantial volume of data and imbuing the model with time-series-related knowledge through pre-training. The second strategy involves fine-tuning based on the LLM framework, wherein corresponding mechanisms are devised to adapt the time series, enabling its application to existing language models. Consequently, this facilitates processing diverse time-series tasks using the pre-existing language models. This path poses challenges and necessitates the ability to transcend the original language model.

A straightforward linear model may have its advantages in specific circumstances; however, it may not be capable of effectively handling extensive time series information on the same level as a more intricate model, such as the transformer. In summary, it is evident that the transformer model remains far from obsolete in time series forecasting. Nonetheless, having abundant training data to fully unlock its immense potential is crucial. Unfortunately, there is currently a scarcity of publicly available datasets that are sufficiently large for time series forecasting. The vast majority of existing pre-trained time-series models are trained using public datasets like Traffic and Electricity. Despite these benchmark datasets serving as the foundation for developing time series forecasting, their limited size and lack of generalizability pose significant challenges for large-scale pre-training. Thus, in the context of time-series prediction, the most pressing matter is the development of expansive and highly generalized datasets (similar to ImageNet in computer vision). This crucial step will undoubtedly propel the advancement of time-series analysis and training models while enhancing the capacity of training models in time-series prediction. Additionally, this development underscores the transformer model's effectiveness in successfully capturing long-term dependencies within a sequence while maintaining superior computational efficiency and a more comprehensive feature representation capability.

## 5 Public datasets and evaluation metrics

In this section, we summarize some common applications and relevant public LTSF datasets. Also, we discuss prediction performance evaluation metrics in LTSF.

### 5.1 Common applications and public datasets

#### 5.1.1 Finance

LTSF is commonly used in finance to predict economic cycles [104], fiscal cycles, and long-term stock trends [105]. In the stock market, LTSF can predict future trends and fluctuations in stock prices [106], helping investors to develop more accurate investment strategies. In financial planning, LTSF can predict future economic conditions, such as income, expenses, and profitability, to help individuals or businesses better plan their financial goals and capital operations [107]. In addition, LTSF can predict a borrower's repayment ability and credit risk [108] or predict future interest rate trends to help financial institutions conduct loan risk assessments for better monetary and interest rate policies. We summarized the open-source LTSF datasets in the finance field in recent years in Table 3.

Table 3. Finance LTSF dataset

| Dataset | Reference | Data information | Min-granularity |
|---|---|---|---|
| Gold prices | [109] | Daily gold prices from 2014.1 to 2018.4, including | 1 day |

| | | minimum, mean, maximum, median, standard deviation, skewness, and kurtosis<br>http://finance.yahoo.com | |
|---|---|---|---|
| GEFCom2014 Electricity Price | [110] | Dataset consists of electricity load forecasting, electricity price forecasting, wind, and solar power generation<br>https://www.dropbox.com/s/pqenrr2mcvl0hk9/GEFCom2014.zip?dl=0 | 1 hour |
| Exchange-rate | [67, 111-113] | Daily exchange rates from Australia, the United Kingdom, Canada, Switzerland, China, Japan, New Zealand, and Singapore between 1990 and 2016<br>https://github.com/laiguokun/multivariate-time-series-data | 1 day |
| S&P 500 | [114] | Daily S&P 500 index from 2001.1 to 2017.5<br>http://finance.yahoo.com | 1 day |
| Shanghai composite | [114] | Daily SSE indices from 2005.1 to 2017.6<br>http://finance.yahoo.com | 1 day |
| S&P 500 stocks | [115] | 505 common stocks traded on the American stock exchange, recording historical daily stock prices for all companies currently included in the S&P 500 index from 2013.2 to 2018.2<br>https://www.kaggle.com/camnugent/sandp500 | 1 day |
| CRSP's stocks | [116] | Data on individual stock returns and prices, S&P 500 index returns, industry categories, number of shares outstanding, ticker symbols, exchange codes, and trading volume<br>http://mba.tuck.dartmouth.edu/pages/faculty/ken.french/data_library.html | 1 day |
| Gas station revenue | [73] | Daily revenue of five gas stations from 2015.12 to 2018.12<br>https://github.com/bighuang624/DSANet/tree/master/data | 1 day |
| Finance Japan | [117] | Dataset collected by the Ministry of Finance of Japan that records general partnerships, limited partnerships, limited liability companies, and joint stock companies from 2003.1 to 2016.12<br>https://www.mof.go.jp/english/pri/reference/ssc/outline.htm | 4 months |
| Stock opening prices | [118] | Daily opening prices for 50 stocks in 10 sectors in Financial Yahoo between 2007 and 2016<br>https://github.com/z331565360/State-Frequency-Memory-stock-prediction | 1 day |

### 5.1.2 Energy

In the energy field, LTSF is often used to assist in developing long-term resource planning strategies [119]. It can help companies and governments forecast future energy demand to better plan energy production and supply. It can also help power companies predict future power generation to ensure a sufficient and stable power supply [120]. In addition, LTSF can help governments and enterprises to develop energy policy planning or manage the energy supply chain [121]. These applications can help enterprises and governments better plan, manage, reduce risks, improve efficiency, and realize sustainable development. We summarized the energy field's open-source datasets in recent years in Table 4.

Table 4. Energy LTSF dataset

| Dataset | Reference | Data information | Min-granularity |
|---|---|---|---|
| Power consumption | [118] | The electricity consumption of a household, including voltage, electricity consumption, and other characteristics from 2006.12 to 2010.11 https://archive.ics.uci.edu/ml/datasets/Individual+household+electric+power+consumption | 1 minute |
| Solar energy | [49, 56, 111, 112, 122] | The highest solar power production from 137 photovoltaic plants in Alabama in 2006 https://www.nrel.gov/grid/solar-power-data.html | 5 minutes |
| electricity | [56, 67, 111, 112, 122] | The electricity consumption of 321 customers between 2011 and 2014 https://archive.ics.uci.edu/ml/datasets/ElectricityLoadDiagrams20112014 | 15 minutes |
| wind | [49, 56] | Hourly estimates of energy potential as a percentage of the maximum output of power plants for a European region between 1986 and 2015 https://www.kaggle.com/sohier/30-years-ofeuropean-wind-generation | 1 hour |
| ETT | [12, 67] | The load and oil temperature of power transformers from 2016.7 to 2018.7 https://github.com/zhouhaoyi/ETDataset | 15 minutes |
| sanyo | [77] | Daily solar power generation data from two photovoltaic plants in Alice Springs, Northern Territory, and Australia from 2011.1 to 2017.1 http://dkasolarcentre.com.au/source/alicesprings/dka-m4-b-phase | 1 day |
| hanergy | [77] | Daily solar power generation data from two photovoltaic plants in Alice Springs, Northern Territory, and Australia from 2011.1 to 2016.12 https://dkasolarcentre.com.au/source/alicesprings/dka-m16-b-phase | 1 day |
| power grid data | [123] | Grid data of State Grid Shanghai Municipal Electric Power Company from 2014.1 to 2015.2 | 1 day |

**5.1.3 Transportation**

In urban transportation, LTSF can help urban traffic management predict future traffic flow [124] for better traffic planning and management. It can also be used to predict future traffic congestion [125], future traffic accident risks, and traffic safety issues [126] for better traffic safety management and accident prevention. We summarized the open-source datasets in the transportation field in recent years in Table 5.

Table 5. Transportation LTSF dataset

| Dataset | Reference | Data information | Min-granularity |
|---|---|---|---|
| Paris metro line | [25] | The passenger flow on Paris metro lines 3 and 13 between 2009 and 2010 http://www.neural-forecasting-competition.com/ | 1 hour |
| PeMS03, PeMS04, PeMS07, PeMS08, | [56, 67, 111, 112, 122, 127-129] | traffic flow data http://pems.dot.ca.gov/ | 30s |
| Birmingham Parking | [115] | The parking lot ID, parking lot capacity, parking lot occupancy, and update time for 30 parking lots operated by Birmingham National Car Park from 2016.10 to 2016.12 http://archive.ics.uci.edu/ml/datasets/parking+birmingham | 30 min |
| METR-LA | [111, 127] | Traffic information collected by loop detectors on Los Angeles County freeways from 2012.3 to 2012.6 https://drive.google.com/drive/folders/10FOTa6HXpQX8Pf5WRoRwcFnW9BrNZEIX | 5 min |
| PEMS-BAY | [111, 127] | Traffic speed readings from 325 sensors collected by PeMS, the California Transit Agency Performance Measurement System from 2017.1 to 2017.5 https://drive.google.com/drive/folders/10FOTa6HXpQX8Pf5WRoRwcFnW9BrNZEIX | 30s |
| SPMD | [130] | The driving records of approximately 3,000 drivers in Ann Arbor, Michigan, from 2015.5 to 2015.10 https://github.com/ElmiSay/DeepFEC | 1 hour |
| VED | [130] | The fuel and energy consumption of various personal vehicles operating under different realistic driving conditions in Michigan, US, from 2017.11 to 2018.11 https://github.com/ElmiSay/DeepFEC | 1 hour |
| England | [128] | National average speeds and traffic volumes derived from UK freeway traffic data from 2014.1 to 2014.6 http://tris.highwaysengland.co.uk/detail/trafficflowdata | 15 min |
| TaxiBJ+ | [131] | The distribution and trajectory of more than 3,000 cabs in Beijing | 30 min |

| Dataset | Reference | Data information | Min-granularity |
|---|---|---|---|
| | | https://www.microsoft.com/enus/research/publication/deep-spatio-temporal-residualnetworks-for-citywide-crowd-flows-prediction | |
| BikeNYC | [131] | Trajectory data taken from the NYC Bike system in 2014, from April 1 to Sept. 30. Trip data includes trip duration, starting and ending station IDs, and start and end times https://www.microsoft.com/enus/research/publication/deep-spatio-temporal-residualnetworks-for-citywide-crowd-flows-prediction | 1 hour |
| HappyValley | [132] | The hourly population density of popular theme parks in Beijing from 2018.1 to 2018.10 heat.qq.com | 1 hour |
| NYC Taxi | [133] | Details of every cab trip in New York City from 2009.1 to 2016.6 | 1 hour |

### 5.1.4 Meteorology and medicine

The application of LTSF in meteorology mainly focuses on predicting long-term climate trends. For example, LTSF can be used to predict long-term climate change [134], providing a scientific basis for national decision-making in response to climate change. It can also issue early warnings for natural climate disasters [135] to mitigate potential hazards to human lives and properties. In addition, LTSF can predict information such as sea surface temperature and marine meteorology for the future [136], providing decision support for industries such as fisheries and marine transportation. We summarized the open-source datasets in the meteorology and medicine fields in recent years in Table 6 and Table 7, respectively.

Table 6. Meteorology LTSF dataset

| Dataset | Reference | Data information | Min-granularity |
|---|---|---|---|
| Beijing PM2.5 | [137] | Hourly PM2.5 data and associated meteorological data for Beijing from 2010.1 to 2014.12 https://archive.ics.uci.edu/ml/datasets.html | 1 hour |
| Hangzhou temperature | [114] | Daily average temperature of Hangzhou from 2011.1 to 2017.1 http://data.cma.cn/data/ | 1 day |
| WTH | [67] | Weather conditions throughout 2020 https://www.bgc-jena.mpg.de/wetter/ | 10 min |
| USHCN | [34] | Continuous daily meteorological records from 1887 to 2009 https://www.ncdc.noaa.gov/ushcn/introduction. | 1 day |
| KDD-CUP | [34] | PM2.5 measurements from 35 monitoring stations in Beijing from 2017.1 to 2017.12 https://www.kdd.org/kdd2018/kdd-cup. | 1 hour |
| US | [129] | Weather datasets from 2012 to 2017 from 36 | 1 hour |

| | | weather stations in the US https://www.kaggle.com/selfishgene/historical-hourly-weather-data. | |

In the medical field, LTSF can be applied to various stages of drug development. For example, predicting a drug's toxicity, pharmacokinetics, pharmacodynamics, and other parameters helps researchers optimize the drug design and screening process [138]. In addition, LTSF can predict medical needs over a certain period [139]. These predictions can be used to allocate and plan medical resources rationally.

Table 7. Medicine LTSF dataset

| Dataset | Deference | Data information | Min-granularity |
|---|---|---|---|
| ILI | [53, 67] | Data on patients with influenza-like illness recorded weekly by the Centers for Disease Control and Prevention from 2002 to 2021 https://gis.cdc.gov/grasp/fluview/fluportaldashboard.html. | 1 week |
| COVID-19 | [140] | Daily data on confirmed and recovered cases collected from 2020.1 to 2020.6 in Italy, Spain, France, China, US, and Australia https://github.com/CSSEGISandData/COVID-19 | 1 day |
| 2020 OhioT1DM | [141] | Eight weeks of continuous glucose monitoring, insulin, physiological sensor, and self-reported life event data for each of 12 patients with type 1 diabetes in 2020 http://smarthealth.cs.ohio.edu/OhioT1DM-dataset.html | 5 min |
| MIMIC-III | [34] | A public clinical dataset with over 58,000 admission records from 2001 to 2012 http://mimic.physionet.org | 1 hour |

### 5.2 Evaluation metrics

In this section, we discuss prediction performance evaluation metrics in the field of TSF. According to [142], the prediction accuracy metrics can be divided into three groups: scale-dependent, scale-independent, and scaled error metrics, which are based on whether the evaluation metrics are affected by the data scale and how the data scale effects are eliminated.

Let $Y_t$ denote the observation at time $t$ ($t = 1,..., n$) and $F_t$ denote the forecast of $Y_t$. Then define the forecast error $e_t = Y_t - F_t$.

### 5.2.1 Scale-dependent measures

Scale-dependent measures are evaluation metrics whose data scales depend on the data size of the original data, and they are the most widely used evaluation metrics in forecasting. These are useful when comparing different methods applied to the same datasets but should not be used, for example, when comparing across datasets with different scales.

The most commonly used scale-dependent measures are based on the absolute error or squared errors:

$$\text{Mean Square Error (MSE)} = \text{mean}(e_t^2) \tag{7}$$
$$\text{Root Mean Square Error (RMSE)} = \sqrt{\text{MSE}} \tag{8}$$
$$\text{Mean Absolute Error (MAE)} = \text{mean}(|e_t|) \tag{9}$$
$$\text{Median Absolute Error (MdAE)} = \text{median}(|e_t|) \tag{10}$$

Historically, the RMSE and MSE have been popular, largely because of their theoretical relevance in statistical modeling. However, they are more sensitive to outliers than MAE or MdAE [143].

**5.2.2 Scale-independent measures**

Scale-independent measures are evaluation metrics not affected by the size of the original data. They can be divided more specifically into three subcategories: measures based on percentage errors, measures based on relative errors, and relative measures.

The percentage error is $p_t = 100e_t/Y_t$. The most commonly used measures are:

$$\text{Mean Absolute Percentage Error (MAPE)} = \text{mean}(|p_t|) \tag{11}$$
$$\text{Median Absolute Percentage Error (MdAPE)} = \text{median}(|p_t|) \tag{12}$$
$$\text{Root Mean Square Percentage Error (RMSPE)} = \sqrt{\text{mean}(p_t^2)} \tag{13}$$
$$\text{Root Median Square Percentage Error (RMdSPE)} = \sqrt{\text{median}(p_t^2)} \tag{14}$$

Percentage errors have the advantage of being scale-independent and so are frequently used to compare forecast performance across different datasets. However, these measures have the disadvantage of being infinite or undefined if $Y_t = 0$ for any *t* in the period of interest and have an extremely skewed distribution when any value of $Y_t$ is close to zero. The MAPE and MdAPE also have the disadvantage of putting a heavier penalty on positive errors than negative errors. Measures based on percentage errors are often highly skewed, and, therefore, transformations (such as logarithms) can make them more stable [144].

An alternative way of scaling is to divide each error by the error obtained using another standard forecasting method. Let $r_t = e_t/e_t^*$ denote the relative error, where $e_t^*$ is the forecast error obtained from the benchmark method. Usually, the benchmark method is the random walk where $F_t$ is equal to the last observation.

$$\text{Mean Relative Absolute Error (MRAE)} = \text{mean}(|r_t|) \tag{15}$$
$$\text{Median Relative Absolute Error (MdRAE)} = \text{median}(|r_t|) \tag{16}$$
$$\text{Geometric Mean Relative Absolute Error (GMRAE)} = \text{gmean}(|r_t|) \tag{17}$$

A serious deficiency of relative error measures is that $e_t^*$ can be small. In fact, $r_t$ has infinite variance because $e_t^*$ has a positive probability density at 0. The use of "winsorizing" can trim extreme values, which will avoid the difficulties associated with small values of $e_t^*$ [145], but adds some complexity to the calculation and a level of arbitrariness as the amount of trimming must be specified.

Rather than use relative errors, one can use relative measures. For example, let $MAE_b$ denote the MAE from the benchmark method. Then, a relative MAE is given by

$$relMAE = MAE/MAE_b. \tag{18}$$

An advantage of these methods is their interpretability. However, they require several forecasts on the same series to compute MAE (or MSE).

**5.2.3 Scaled errors**

Scaled errors were first proposed in [142] and can be used to eliminate the effect of data size by comparing the prediction results obtained with the underlying method (usually the native method). The following scaled error is commonly used:

$$q_t = \frac{e_t}{\frac{1}{n-1}\sum_{i=2}^{n}|Y_i - Y_{i-1}|}. \tag{19}$$

Therefore, The Mean Absolute Scaled Error is simply $MASE = \text{mean}(|q_t|)$.

There is a simple way to understand this evaluation metric; the denominator can be understood as the average error of the native predictions made one step ahead in the future. If the MASE > 1, then the experimental method under evaluation is worse than the native prediction, and vice versa. While MdASE is similar to MASE, the way MASE is calculated using the mean makes it more susceptible to outliers, while MdASE calculated using the median has stronger robustness and validity. However, such metrics can only reflect the results of comparison with the basic method and cannot visualize the error of the prediction results.

**6 Training strategies**

Recent transformer variants introduce various time-series features into the models for improvements [67, 70]. In this section, we summarize several training strategies of existing transformer-based models for LTSF.

**6.1 Preprocessing and embedding**

In the preprocessing stage, normalization with zero mean is often applied in time-series tasks. Moreover, seasonal-trend decomposition is a standard method to make raw data more predictable [146, 147], first proposed by Autoformer [67]. It also uses a moving average kernel on the input sequence to extract the trend-cyclical component of the time series. The seasonal component is the difference between the original sequence and the trend component. FEDformer [79] further proposes a mixture of experts' strategies to mix the trend components extracted by moving average kernels with various kernel sizes.

The self-attentive layer in the transformer architecture cannot preserve the positional information of the time series. However, local location information or the ordering of the time series is essential. Furthermore, global time information, such as hierarchical timestamps (weeks, months, years) and agnostic timestamps (holidays and events), is also informative [12]. To enhance the temporal context of the time-series input, a practical design is to inject multiple embeddings into the input sequence, such as fixed positional coding and learnable temporal embeddings. Additionally, the introduction of temporal embeddings accompanied by temporal convolutional layers [49] or learnable timestamps [67] has been proposed as effective means to further enhance the temporal context of the input data..

**6.2 Iterated multi-step and direct multi-step**

The time series forecasting task is to predict the values at the $T$ future time steps. When $T > 1$, iterated multi-step (IMS) forecasting [148] learns a single-step forecaster and iteratively applies it to obtain multi-step predictions. Alternatively, direct multi-step (DMS) forecasting [149] optimizes the multi-step forecasting objective simultaneously. The variance of the IMS predictions is smaller due to the autoregressive estimation procedure compared to DMS forecasting but is inevitably subject to the error accumulation effects. Therefore, IMS forecasting is more desirable when highly accurate single-step forecasters exist, and $T$ is relatively small. In contrast, DMS forecasting produces more accurate forecasts when unbiased single-step forecast models are challenging to obtain or when $T$ is large.

Applying the vanilla transformer model to the LTSF problem has some limitations, including the quadratic time/memory complexity with the original self-attention scheme and error accumulation caused by the autoregressive decoder design. Alternative transformer variants have been developed to overcome these challenges, each employing distinct strategies. For instance, LogTrans [49] introduces a dedicated decoder for IMS forecasting, while Informer [12] leverages a generative-style decoder. Additionally, Pyraformer [70] incorporates a fully connected layer that concatenates spatiotemporal axes as its decoder. Autoformer [67] adds the two refined decomposition features of the trend-cyclical components and the stacked autocorrelation mechanism of the seasonal component to obtain the final prediction results. Similarly, FEDformer [79] applies a decomposition scheme and employs the proposed frequency attention block in deciphering the final results.

**7 Conclusion**

Transformer architecture has found its application in solving various time-series tasks. The transformer architecture based on self-attention and positional encoding offers better or similar performance as RNNs and variants of LSTMs/GRUs. However, it is more efficient in computing time and overcomes other shortcomings of RNNs/LSTMs/GRUs.

In this paper, we summarized the application of the transformer on LTSF. First, we have provided a thorough examination of the fundamental structure of the transformer. Subsequently, we analyzed and summarized the advantages of transformer on LTSF tasks. Given that the transformer encounters intricacies and interdependencies when confronting LTSF tasks, numerous adaptations have been introduced to the original architectural framework, thus equipping transformers with the capacity to handle LTSF tasks effectively. This architectural augmentation, however, brings certain challenges during the training process. To address this, we have incorporated a compendium of best practices that facilitate the practical training of transformers. Additionally, we have collected abundant resources on TSF and LTSF, including datasets, application fields, and evaluation metrics.

In summary, our comprehensive review presents an in-depth examination of recent advancements in tranformer-based architecture in LTSF and imparts valuable insights to researchers seeking to improve their models. The transformer architecture is renowned for its remarkable modeling capacity and aptitude for capturing long-term dependencies. However, it encounters challenges regarding time complexity when applied to LTSF tasks. While efforts to reduce complexity may inadvertently lead to the loss of certain interdependencies between data points, thereby compromising prediction accuracy. Consequently, the amalgamation of various techniques within a compound model, leveraging the strengths of each, emerges as a promising avenue for future research in transformer-based LTSF models. This paves the way for innovative model designs, data processing techniques, and benchmarking approaches to tackle the intricate LTSF problems. Notably, researchers have recently explored the integration of Large Language Models (LLMs) in time series forecasting, wherein LLMs exhibit the capability to generate forecasts while offering human-readable explanations for predictions, outperforming traditional statistical models and machine learning approaches. These encouraging findings present a compelling impetus for further exploration, aiming to enhance the precision, comprehensibility, and transparency of forecasting results.

**Conflict of interest**

The authors declare that they have no conflict of interest.

**Data availability**

Data sharing not applicable to this article as no datasets were generated or analyzed during the current study.

**Acknowledgment**

This work was supported by the Project of the Educational Commission of Guangdong Province of China (2022ZDJS113), and the Natural Science Foundation of Top Talent of SZTU (GDRC20221).